\documentclass[twoside,11pt]{article}

\usepackage{blindtext}

\usepackage{jmlr2e}

\usepackage{lastpage}
\jmlrheading{23}{2026}{1-\pageref{LastPage}}{1/21; Revised 5/22}{9/22}{21-0000}{Quentin Nater and Mourad Khayati}

\ShortHeadings{Sample JMLR Paper}{A Comprehensive Library for Time Series Imputation}
\firstpageno{1}

\usepackage{balance}
\usepackage[utf8]{inputenc}
\usepackage[T1]{fontenc}
\usepackage{url}
\usepackage{mathtools}
\usepackage{diagbox}
\usepackage{multirow, booktabs}
\usepackage{hyperref}
\usepackage{colortbl}
\usepackage{amsmath}
\usepackage{amsfonts}
\usepackage{verbatim} 
\usepackage[boxed]{algorithm} 
\usepackage{algpseudocode}
\usepackage[algo2e,titlenotnumbered,boxed,ruled,vlined,linesnumbered]{algorithm2e}
\SetAlFnt{\small\sffamily}

\usepackage{dcolumn}
\usepackage{booktabs}
\usepackage{lscape}
\usepackage{listings}

\usepackage{flexisym}
\usepackage[normalem]{ulem}
\usepackage{caption}
\usepackage{subcaption}

\usepackage{tikz,pgfplots,pgfplotstable}
\pgfplotsset{compat=1.8}
\usetikzlibrary{bayesnet}
\usetikzlibrary{pgfplots.groupplots}

\usetikzlibrary{patterns}
\usetikzlibrary{decorations.pathreplacing}
\usetikzlibrary{matrix,positioning,decorations.pathreplacing}
\usetikzlibrary{arrows}
\usetikzlibrary{3d,calc}
\usetikzlibrary{arrows.meta,hobby}

\usepgfplotslibrary{statistics}

\usepackage{adjustbox}
\usepackage{rotate}

\usepackage{enumitem}
\usepackage{siunitx}
\usepackage{graphicx}
\graphicspath{{../}{../figs/}}
\usepackage{epstopdf}
\usepackage{cleveref}
\usepackage[a-2b]{pdfx}

\definecolor{mauve}{rgb}{0.58, 0.0, 0.83}

\newcommand{\quotes}[1]{``#1''}
\newcommand{\sys}{ImputeGAP\xspace} 
\newcommand{\softbsubsec}[1]{\vspace{0.5em}\noindent\textbf{#1.}}
\newcommand{\softbsubsecR}[1]{\vspace{0.48em}\noindent\textbf{#1.}}

\usepackage{wasysym}
\usepackage{pifont}
\newcommand{\xmark}{\ding{53}}%

\begin{document}

\title{\sys: A Comprehensive Library for\\ Time Series Imputation}

\author{\name Quentin Nater \email quentin.nater@unifr.ch \\
       \name Mourad Khayati \email mourad.khayati@unifr.ch \\
       \addr Department of Computer Science\\
       University of Fribourg\\
       Boulevard de Pérolles 90, 1700 Fribourg, Switzerland
       }

\editor{My editor}

\maketitle

\begin{abstract}

With the prevalence of sensor failures, imputation, the process of estimating missing values, has emerged as the cornerstone of time series data pre-processing. While numerous imputation algorithms have been developed to repair these data gaps, existing time series libraries provide limited imputation support. Furthermore, they often lack the ability to simulate realistic time series missingness patterns and fail to account for the impact of the imputed data on subsequent downstream analysis.

This paper introduces \sys, a comprehensive library for time series imputation that supports a diverse range of imputation methods and modular missing data simulation, catering to datasets with varying characteristics. The library includes extensive customization options, such as automated hyperparameter tuning, benchmarking, explainability, downstream evaluation, and compatibility with popular time series frameworks.

\end{abstract}


\section{Introduction}

The proliferation of the Internet of Things (IoT) has propelled the seamless integration of sensors into our daily lives. From smart grids and health monitor systems to automated vehicles and urban traffic systems, sensors generate unprecedented volumes of time series data. One critical issue with this data deluge is the rising occurrence of temporary data transfer failures due to factors such as network disruptions, hardware malfunctions, and environmental interference. These interruptions, though often brief, can lead to gaps of consecutive values in the collected time series data. Such quality issues can undermine the reliability and integrity of data, particularly in downstream applications
such as predictive analytics, similarity search, and real-time monitoring systems.

The diversity in time series characteristics and missingness data patterns has led to the development of various families of imputation algorithms designed to address data gaps. These techniques aim to generate plausible estimates for missing segments, offering a wide range of accuracy and efficiency trade-offs. Over the past decade, numerous libraries and frameworks have been introduced to streamline the development and deployment of imputation algorithms, facilitating their integration into practical workflows.

\softbsubsec{Imputation Libraries} Imputation libraries can be broadly classified into two distinct categories based on the type of data they handle. The first category comprises generic libraries designed for tabular data, such as gcimpute~\citep{DBLP:journals/jstatsoft/ZhaoU24}, miceforest~\citep{10.1093/aje/kwt312}, autoimpute~\citep{autoimpute}, fancyimpute~\citep{fancyimpute}, scikit-learn~\citep{sklearn_api}, and others. While these libraries are easy to deploy, they fall short in incorporating the essential temporal structure of time series data in both the imputation algorithms and missingness patterns.

The second category consists of specialized libraries designed for time series data. Table~\ref{tab:library-comparison} presents a comparative analysis of these solutions, showcasing how \sys advances the field. The first two columns of the table \quotes{Contamination} and \quotes{Imputation Family}, outline the type of missing data they simulate, ‘Mono-block’ introduces a single missing block per series with variable size and position and ‘Multi-block’ inserts multiple missing blocks per series with variable numbers and positions, as well as the algorithm families they implement, including Statistical, Machine Learning, Pattern Search, Matrix Completion, Deep Learning, and Large Language Models. The last two columns, \quotes{Imputation Explanation} and \quotes{Downstream Analysis}, highlight whether the libraries provide interpretation of the algorithm's behavior and assess the imputation impact on downstream tasks.

\begin{table*}[htb!]
    \centering
    \caption{A comparison of available time series imputation libraries. The symbols indicate the following: {\checkmark} for present, {(\checkmark)} for
    incomplete, and {\xmark} for absent.}
    \label{tab:library-comparison}
    \resizebox{\textwidth}{!}
    {
        \begin{tabular}{|c|c|c|c|c|c|c|c|c|c|c|}
        \hline
        Library & \multicolumn{2}{c|}{Contamination} & \multicolumn{6}{c|}{Imputation Family} & Imputation & Downstream \\
        \cline{2-9}
         & Mono-block & Multi-block & Stats & ML & PSearch & MatComp & DL & LLM & Explanation  & Analysis  \\ \hline

        cleanTS~\citep{DBLP:journals/ijon/ShendeFB22} & $(\checkmark)$  & \xmark  & $\checkmark$ & \xmark & \xmark & \xmark & \xmark & \xmark & \xmark & \xmark \\ \hline
        
        sktime~\citep{DBLP:journals/corr/abs-1909-07872} & \xmark  & \xmark  & $\checkmark$ & \xmark & \xmark & \xmark & \xmark & \xmark & \xmark & \xmark  \\ \hline
    
        amelia II~\citep{JSSv045i07} & \xmark  & \xmark  & $\checkmark$ & $(\checkmark)$ & \xmark & \xmark & \xmark & \xmark & \xmark & \xmark \\ \hline
        
        PyPOTS~\citep{DBLP:journals/corr/abs-2305-18811} & $(\checkmark)$  & $(\checkmark)$  & $\checkmark$ & \xmark & \xmark & \xmark & \checkmark & \checkmark & \xmark & \xmark \\ \hline \hline
    
        ImputeGAP & $\checkmark$ & $\checkmark$ & $\checkmark$ & $\checkmark$ & $\checkmark$ &  $\checkmark$ & $\checkmark$ & $\checkmark$ & $\checkmark$   & $\checkmark$\\ \hline
        
        \end{tabular}
   }
\end{table*}

\softbsubsec{Contributions} In this work, we present \sys, a comprehensive library designed to overcome the limitations of existing time series imputation frameworks. Our library distinguishes itself by offering a diverse range of advanced imputation algorithms, along with a configurable contamination module that simulates real-world missingness patterns. Additionally, \sys includes tools to analyze the behavior of these algorithms and assess their impact on key downstream tasks in time series analysis, such as forecasting. The library is available on PyPI through \url{https://pypi.org/project/imputegap/}.

\section{\sys}
\label{sec:imputegap}

\subsection{Architecture}
\label{subsec:arch}

\begin{figure}[h!]
  \centering
  \resizebox{0.85\columnwidth}{!}{%
    \includegraphics{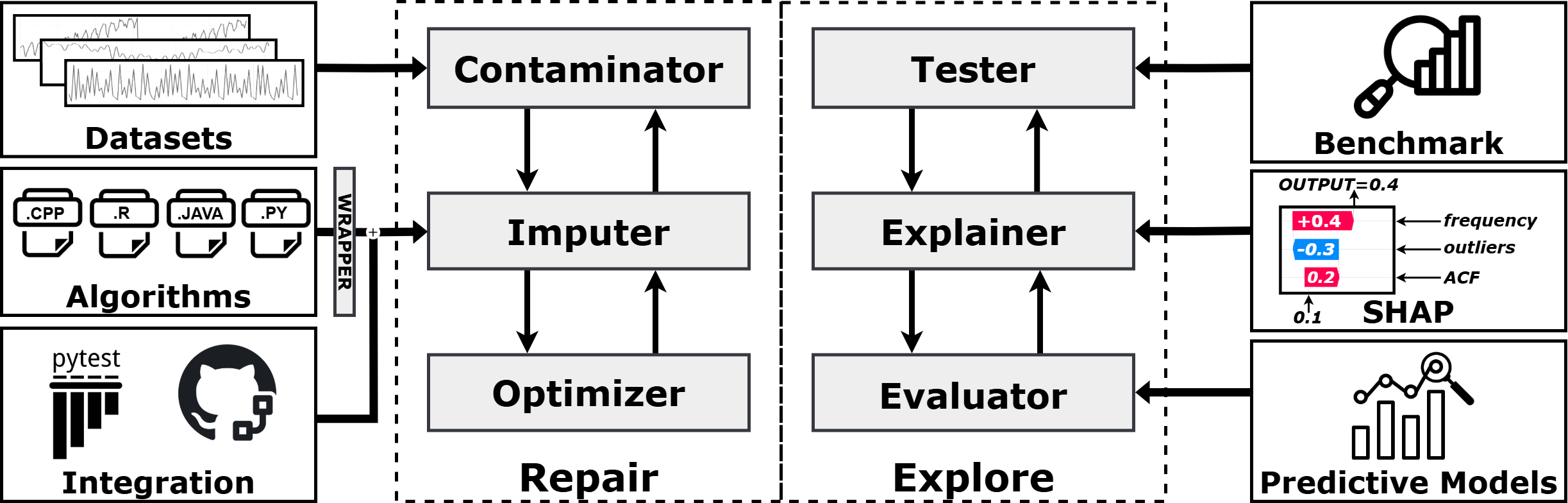}
  }
  \caption{The \sys Framework.}
  \label{fig:sys_overview}
\end{figure}

\sys is an end-to-end imputation library that implements the full imputation pipeline from data collection to explaining the imputation results and their impact. It encompasses two interleaving units: repair and explore. 
The two units can be accessed via a standardized pipeline defined by configuration files or independent instantiation.
This modular design provides a unified platform for upstream and downstream imputation evaluation. 
Figure~\ref{fig:sys_overview} represents the library's architecture, highlighting the components contributing to the imputation evaluation process.

\subsection{Modules}

\softbsubsec{Contaminator (GenGap)} This component has two primary functions: loading the data and simulating missingness patterns. 
By instantiating the Time Series object, the contaminator populates the necessary classes, ensuring they interact deterministically.
To load the data, \sys provides access to a diverse collection of time series datasets, including those from popular imputation benchmarks~\citep{DBLP:journals/pvldb/KhayatiLTC20,DBLP:journals/tkde/MiaoWCGY23, DBLP:journals/corr/abs-2406-12747}, while also allowing user-supplied datasets. Users have full control over data contamination, introducing one or multiple missing blocks per series. In the mono-block case, they can adjust the contamination rate, from 1\% to 80\% of the series length, and position it to create overlapping, disjoint, or blackout patterns. For multi-block scenarios, users can indicate the size of missing blocks per series, the contamination rate, and determine their placement, whether randomly assigned or following a specific distribution. In both cases, users can indicate the percentage of time series per dataset that is contaminated.

\begin{lstlisting}[
    language=Python, 
    label={listing:contamination}
]
from imputegap.recovery.manager import TimeSeries
from imputegap.recovery.contamination import GenGap
ts = TimeSeries()
ts.load_series(utils.search_path("eeg-alcohol"))
ts_m = GenGap.mcar(ts.data, rate_dataset=0.2, rate_series=0.4, block_size=10)
\end{lstlisting}

\softbsubsec{Imputer} This is the central component that triggers the imputation workflow within the framework.
Once the imputation object is created, the user can execute the imputation process
using either the algorithm’s predefined default parameters or by specifying custom parameters
in a dictionary. The resulting imputation is stored as a matrix within the object and can be passed to
the scoring function, which compares the imputation results against their ground truth. The Imputer offers access to a wide range of ready-to-deploy algorithms. Alternatively, users may integrate their algorithm in various languages such as Python, C++, Java, and R.

\begin{lstlisting}[
    language=Python, 
    label={listing:imputation}
]
from imputegap.recovery.imputation import Imputation
imp = Imputation.MatrixCompletion.CDRec(ts_m)
imp.impute(params={"rank": 5, "epsilon": 0.01, "iterations": 100})
imp.score(ts.data, imp.recov_data)
\end{lstlisting}

\softbsubsec{Optimizer} The Optimizer component manages algorithms' configuration and hyperparameter tuning. To invoke the tuning process, users need to specify the optimization option during the Impute call by selecting the appropriate input for the algorithm. The parameters are defined by providing a dictionary containing the ground truth, the chosen optimizer, and the optimizer's options. Several search algorithms are available, including those provided by Ray Tune~\citep{liaw2018tune}.

\begin{lstlisting}[
    language=Python, 
    label={listing:automl}
]
params = {"input_data": ts.data, "optimizer": "ray_tune"}
imp.impute(user_def=False, params=params)
\end{lstlisting}

\softbsubsecR{Tester} The library can serve as a test-bed for comparing the performance of imputation 
algorithms. It provides a suite of benchmarking tools and customized plot generation that leverage the Imputer module. Users may specify a subset of algorithms for comparison
and select the imputation metric for their analysis. \sys implements various evaluation metrics, each capturing a different aspect of the imputation quality.

\begin{lstlisting}[
    language=Python, 
    label={listing:benchmark}
]
from imputegap.recovery.benchmark import Benchmark
bench = Benchmark()
bench.eval(algorithms=["CDRec", "TimesNet"], datasets=["eeg-alcohol"], patterns=["mcar", "aligned"], metrics=["RMSE", "MI"])
\end{lstlisting}

\softbsubsecR{Explainer} One of the salient features of \sys is to provide insights into the algorithm's behavior.
By training a regression model to predict imputation results across various methods, \sys leverages SHapley Additive exPlanations, SHAP\citep{NIPS2017_7062}, to reveal how different time series features influence the model’s predictions. The library interfaces with various feature extractors, such as Catch22~\citep{Lubba2019}, TSFresh~\citep{CHRIST201872}, and TSFEL~\citep{BARANDAS2020100456}.

\begin{lstlisting}[
    language=Python, 
    label={listing:explainer}
]
from imputegap.recovery.explainer import Explainer
exp = Explainer()
exp.shap_explainer(ts.data, algorithm="cdrec", extractor="pycatch")
\end{lstlisting}

\softbsubsec{Evaluator} The downstream evaluator complements the Impute component with 
a collection of models to assess the impact of imputation on downstream analytics. The library supports a variety of forecasters 
trained on the imputed time series and evaluates their accuracy in predicting future values based on past data.
\sys provides a suite of tools to evaluate the output of downstream models across the imputation algorithms.

\begin{lstlisting}[
    language=Python, 
    label={listing:downstream}
]
imp.score(ts.data, imp.recov_data, downstream={"task": "forecast", "model": "prophet", "baseline": "cdrec"}
\end{lstlisting}

\section{Conclusion}
\label{sec:conc}

In this paper, we introduced \sys, a comprehensive library for imputation algorithms. \sys stands apart from existing libraries by integrating a wide variety of imputation algorithm families and addressing diverse patterns of missingness. Moreover, the library offers innovative features, including robust benchmarking capabilities, tools for explainability of imputation results, and the ability to evaluate the impact of imputation on downstream analytical tasks.




\bibliography{main.bib}

\end{document}